\pdfoutput=1

\documentclass[11pt]{article}

\usepackage{acl}

\usepackage{times}
\usepackage{latexsym}

\usepackage[T1]{fontenc}

\usepackage[utf8]{inputenc}

\usepackage{microtype}

\usepackage{graphicx}
\usepackage{tabularx}
\usepackage{adjustbox}
\usepackage{booktabs}
\usepackage{multirow}
\usepackage{amsmath}
\usepackage{cuted}
\usepackage{comment}
\usepackage{lipsum}
\usepackage{colortbl}
\usepackage{xcolor}

\title{When do Contrastive Word Alignments Improve \\ Many-to-many Neural Machine Translation?}

\author{
Zhuoyuan Mao $^{\mu}$ \hspace{1em}
Chenhui Chu $^\mu$ \hspace{1em} 
Raj Dabre $^\nu$ \hspace{1em} \\
{\bf Haiyue Song $^\mu$} \hspace{1em}
{\bf Zhen Wan $^\mu$} \hspace{1em}
{\bf Sadao Kurohashi $^\mu$}\\
$^\mu$ Kyoto University, Japan \hspace{1em}
$^\nu$ NICT, Japan \\
\texttt{\{zhuoyuanmao, chu, song, zhenwan, kuro\}@nlp.ist.i.kyoto-u.ac.jp} \\
\texttt{raj.dabre@nict.go.jp}
}

\begin{document}
\maketitle
\begin{abstract}

Word alignment has proven to benefit many-to-many neural machine translation (NMT). However, high-quality ground-truth bilingual dictionaries were used for pre-editing in previous methods, which are unavailable for most language pairs. Meanwhile, the contrastive objective can implicitly utilize automatically learned word alignment, which has not been explored in many-to-many NMT. This work proposes a word-level contrastive objective to leverage word alignments for many-to-many NMT. Empirical results show that this leads to 0.8 BLEU gains for several language pairs. Analyses reveal that in many-to-many NMT, the encoder's sentence retrieval performance highly correlates with the translation quality, which explains when the proposed method impacts translation. This motivates future exploration for many-to-many NMT to improve the encoder's sentence retrieval performance.




\end{abstract}

\section{Introduction}
Many-to-many neural machine translation (NMT)~\cite{firat-etal-2016-multi,johnson-etal-2017-googles,aharoni-etal-2019-massively,sen-etal-2019-multilingual,DBLP:journals/corr/abs-1907-05019,lin-etal-2020-pre,pan-etal-2021-contrastive} jointly trains a translation system for multiple language pairs and obtain significant gains consistently across many translation directions. Previous work~\cite{lin-etal-2020-pre} shows that word alignment information helps improve pre-training for many-to-many NMT. However, manually cleaned high-quality ground-truth bilingual dictionaries are used to pre-edit the source sentences, which are unavailable for most language pairs.

Recently, contrastive objectives~\cite{DBLP:conf/iclr/ClarkLLM20,DBLP:conf/iclr/GunelDCS21,DBLP:conf/acl/GiorgiNWB20,DBLP:conf/iclr/WeiW0XYL21,mao-etal-2021-lightweight} have been shown to be superior at leveraging alignment knowledge in various NLP tasks by contrasting the representations of positive and negative samples in a discriminative manner. This objective, which should be able to utilize word alignment learned by any toolkit, which in turn will remove the constraints of using manually constructed dictionaries, has not been explored in the context of leveraging word alignment for many-to-many NMT.

An existing contrastive method~\cite{pan-etal-2021-contrastive} for many-to-many NMT relies on sentence-level alignments. Given that the incorporation of word alignments has led to improvements in previous work, we believe that fine-grained contrastive objectives focusing on word alignments should help improve translation. Therefore, this paper proposes word-level contrastive learning for many-to-many NMT using the word alignment extracted by automatic aligners. We conduct experiments on three many-to-many NMT systems covering general and spoken language domains. Results show that our proposed method achieves significant gains of 0.8 BLEU in the general domain compared to previous word alignment based methods and the sentence-level contrastive method.

We then analyze how the word-level contrastive objective affects NMT training. Inspired by previous work~\cite{artetxe-schwenk-2019-massively} that train sentence retrieval models using many-to-many NMT, we speculate that our contrastive objectives affect the sentence retrieval performance and subsequently impact the translation quality. Further investigation reveals that in many-to-many NMT, the sentence retrieval precision of the multilingual encoder for a language pair  strongly correlates with its translation quality (BLEU), which provides insight about when contrastive alignment improves translation. This revelation emphasizes the importance of improving the retrieval performance of the encoder for many-to-many NMT.

\section{Word-level Contrastive Learning for Many-to-many NMT}

Inspired by the contrastive learning framework~\cite{DBLP:conf/icml/ChenK0H20} and the sentence-level contrastive learning objective~\cite{pan-etal-2021-contrastive}, we propose a word-level contrastive learning objective to explicitly guide the training of the multilingual encoder to obtain well-aligned cross-lingual representations. Specifically, we use word alignments, obtained using automatic word aligners, to supervise the training of the multilingual encoder by a contrastive objective alongside the NMT objective.

\noindent{\textbf{Alignment Extraction}} 
\label{alignex}
Two main approaches for automatically extracting aligned words from a sentence pair are: using a bilingual dictionary and using unsupervised word aligners. The former extracts fewer but precise alignments, whereas the latter extracts more but noisy alignments. We extract word-level alignments by both methods and explore how they impact NMT training. For the former approach, we use word2word~\cite{choe-etal-2020-word2word} to construct bilingual lexicons and then extract word pairs from parallel sentences. The extracted word pairs are combined to form a phrase if words are consecutive in the source and target sentence. For the latter approach, we use FastAlign~\cite{dyer-etal-2013-simple} and use only 1-to-1 mappings for training.

\noindent{\textbf{Word-level Contrastive Learning}}
With the extracted alignments, we propose a word-level contrastive learning objective for the multilingual encoder by the motivation that the aligned words within a sentence pair should have a similar contextual representation. We expect the supervision of the contrastive objective on the corresponding contextual word representation leads to a robust multilingual encoder. Assume that the tokenized source and target parallel sentences in the $i-th$ batch are $\mathcal{D}_i=\{{src}_{ij}, {tgt}_{ij}\}_{j=1}^B$, and the extracted alignments from all the sentence pairs in each batch are $\mathcal{A}_{i}=\{{s}_{ik}, {t}_{ik}\}_{k=1}^N$, where $B$ and $N$ denote the batch-size and the number of alignments, respectively. Note that ${s}_{ik}$ and ${t}_{ik}$ may contain several tokens after the word combination for word2word or subword tokenization for NMT. Then the word-level contrastive loss in a batch is:
\begin{equation}
\resizebox{\linewidth}{!}{
$
\begin{aligned}
\mathcal{L}_{align}^{(i)} =
-\sum_{k=1}^N(\log\frac{\exp{(sim(s_{ik}, t_{ik}) / \mathcal{T})}}{\sum_{m=1}^N \exp{(sim(s_{ik}, t_{im}) / \mathcal{T})}} \\ + \log\frac{\exp{(sim(s_{ik}, t_{ik}) / \mathcal{T})}}{\sum_{m=1}^N \exp{(sim(s_{im}, t_{ik}) / \mathcal{T})}} )
\end{aligned}
$
}
\label{align}
\end{equation}
where $\mathcal{T}$ denotes a similarity scaling temperature. The similarity between two words is measured by:
\begin{align}
sim(word_x,word_y) &= cos(g(\mathbf{\Bar{x}}), g(\mathbf{\Bar{y}}))
\label{sim}
\end{align}
where $g(\mathbf{x}) = \mathbf{W}_2\sigma(\mathbf{W}_1\mathbf{x})$ and $\mathbf{\Bar{x}}$ denotes the average of contextual hidden states of the corresponding subword positions on top of the multilingual encoder. Following~\cite{DBLP:conf/icml/ChenK0H20}, we use an MLP between contrastive loss and the contextual representation for NMT loss. ReLU activation is used for $\sigma$, $\mathbf{W}_1$ is $d\times d$ and $\mathbf{W}_2$ is $d\times d'$, where $d$ is the encoder's hidden dimension and $d'<d$ .

Finally, to jointly train with the NMT loss, we use the following equation to combine our proposed word-level contrastive loss for a batch:
\begin{equation}
\mathcal{L}^{(i)} = \frac{1}{B}(\mathcal{L}_{NMT}^{(i)} + w\frac{N_T}{2N}\mathcal{L}_{align}^{(i)})
\end{equation}
where $N_T$ is the number of the tokens within a batch, $\frac{N_T}{2N}$ is a multiplier that scales the contrastive loss to be consistent with NMT loss, and $w$ is a weight to balance the joint training.

\begin{table}[t]
    \centering
    \resizebox{\linewidth}{!}{
    \begin{tabular}{llr|rr}
    \toprule
        La. pair & Source & Size & N (w2w) & N (FA) \\
         \hline
         en-et & WMT18 & 1.9M & 5,762,977 & 38,454,477 \\
         en-it & IWSLT17 & 231k & 603,032 & 3,000,011 \\
         en-ja & IWSLT17 & 223k & 684,583 & 2,797,882 \\
         en-kk & WMT19 & 124k & 124,511 & 279,429 \\
         en-my & ALT & 18k & 75,383 & 377,392 \\
         en-nl & IWSLT17 & 237k & 564,697 & 2,836,873 \\
         en-ro & WMT16 & 612k & 3,271,848 & 13,092,240 \\
         en-tr & WMT17 & 207k & 770,873 & 2,885,102 \\
         en-vi & IWSLT15 & 133k & 354,167 & 2,120,755 \\
         \bottomrule
    \end{tabular}
    }
    \caption{\textbf{Data Source and number of the extracted word pairs.} La. pair, N (w2w) and N (FA) denote the language pair, the number of the word pairs extracted by word2word and FastAlign, respectively. Refer to Appendix~\ref{sec:app2} for details of the dataset splits.}
    \label{tab1}
\end{table}

\section{Experimental Settings}
\begin{table*}[t]
    \centering
    \resizebox{0.8\linewidth}{!}{
    \begin{tabular}{lrrrrrrrrrr}
    \toprule
        \multirow{2}{*}{Methods} & \multicolumn{2}{c}{en-tr} & \multicolumn{2}{c}{en-ro} & \multicolumn{2}{c}{en-et} & \multicolumn{2}{c}{en-kk} & \multicolumn{2}{c}{en-my} \\
        &$\rightarrow$&$\leftarrow$&$\rightarrow$&$\leftarrow$&$\rightarrow$&$\leftarrow$&$\rightarrow$&$\leftarrow$&$\rightarrow$&$\leftarrow$\\
        \toprule
        MLSC & 9.3&	12.6&	25.0&	26.2&	10.8&	15.1 & 0.5	&5.3	&15.1	&15.6\\
        \hline
        \ \ +align & 9.0&12.4&24.6&\cellcolor{cyan!25}26.5&10.7&14.6& 0.4 & \cellcolor{cyan!25}5.4 & 15.0 & 15.3\\
        \ \ +w2w (ours) &\cellcolor{cyan!25}9.4&	12.6&	24.8&	\cellcolor{cyan!80}26.8&	10.8&	15.1	&0.5&
        \cellcolor{cyan!80}5.8&	\cellcolor{cyan!25}15.2&	\cellcolor{cyan!25}15.9\\
        \ \ +FA (ours) & 9.1&	12.2&	24.8&	\cellcolor{cyan!80}26.7&	10.7&	14.8& 	0.3&	\cellcolor{cyan!25}5.6&	15.0&	15.6	\\
        \toprule
        mBART FT & 17.7&	22.2&	33.8&	37.1&	14.5&	24.3 &	1.8&	14.1&	17.8&	23.1\\
        \hline
        \ \ +align & 17.5&21.9&33.8&36.7&\cellcolor{cyan!80}15.2&24.3 & 1.8&14.0&16.9&22.1 \\
        \ \ +w2w (ours) &17.6&	22.2&	\cellcolor{cyan!25}34.2&	\cellcolor{cyan!25}37.5&	\cellcolor{cyan!80}15.0&	\cellcolor{cyan!80}25.0&1.2&	14.1&	\cellcolor{cyan!80}18.3&	\cellcolor{cyan!80}23.8\\
        \ \ +FA (ours) & 17.5&	22.2&	\cellcolor{cyan!80}34.3&	\cellcolor{cyan!25}37.5	&\cellcolor{cyan!25}14.9&	\cellcolor{cyan!80}25.1& 1.3&	\cellcolor{cyan!25}14.4&	\cellcolor{cyan!25}17.9&	\cellcolor{cyan!80}23.6\\
        \bottomrule
    \end{tabular}
    }
    \caption{\textbf{BLEU scores of 626\_en-tr-ro-et-my-kk system.} Significantly better scores~\cite{koehn-2004-statistical} are in cyan, and marginal improvements are in lightcyan.}
    \label{bleu2}
\end{table*}

\noindent{\textbf{Datasets and Preprocessing}}
We selected ten languages, including English (en), Estonian (et), Italian (it), Japanese (ja), Kazakh (kk), Burmese (my), Dutch (nl), Romanian (ro), Turkish (tr), Vietnamese (vi) from different language families to train the NMT systems. We used the parallel datasets from different domains for the selected nine language pairs, including IWSLT, WMT, and ALT. We followed mBART~\cite{liu-etal-2020-multilingual-denoising} for tokenization. Details are given in Appendix~\ref{sec:app1}. For each parallel dataset, we implemented two approaches as stated in Section~\ref{alignex} to extract word pairs for the contrastive training objective. Data source and the number of the extracted word pairs are shown in Table~\ref{tab1}. To ensure high alignment quality, we used large-scale out-of-domain (see Appendix~\ref{sec:app2}) parallel corpora with FastAlign.

\begin{table}[t]
    \centering
    \resizebox{0.85\linewidth}{!}{
    \begin{tabular}{lrrr}
    \toprule
        Methods & 222\_en-ja & 626\_I & 626\_II \\
        \toprule
        MLSC & 13.90 & 23.76 & 13.55 \\
        \hline
        \ \ +align & 13.90 & 23.67 & 13.39 \\
        \ \ +w2w (ours) & 13.85 & 23.44 & \textbf{13.69} \\
        \ \ +FA (ours) & 13.30 & 23.68 & 13.48 \\
        \toprule
        mBART FT & 18.90 & 29.11 & 20.64 \\
        \hline
        \ \ +align & 18.55 & 28.87 & 20.42 \\
        \ \ +w2w (ours) & 18.80 & 29.08 & \textbf{20.89} \\
        \ \ +FA (ours) & 18.65 & 29.01 & \textbf{20.87} \\
        \bottomrule
    \end{tabular}
    }
    \caption{\textbf{Overall average BLEU of all the systems.} 626\_I and 626\_II denote 626\_en-it-ja-nl-tr-vi and 626\_en-tr-ro-et-my-kk, respectively. Results better than MLSC or mBART FT are marked \textbf{bold}. Refer to Appendix~\ref{sec:app3} for the detailed scores of all the systems.}
    \label{bleu1}
\end{table}

\noindent{\textbf{Many-to-many NMT systems}}
We established three many-to-many NMT systems as follows:

\textbf{222\_en-ja}: Bidirectional en-ja NMT model using en-ja parallel corpus.
    
\textbf{626\_en-it-ja-nl-tr-vi}: 6-to-6 multilingual NMT model using spoken language domain corpora for en-it, en-ja, en-nl, en-tr and en-vi.

\textbf{626\_en-tr-ro-et-my-kk}: 6-to-6 multilingual NMT model using general domain corpora for en-tr, en-ro, en-et, en-my and en-kk.

\noindent{\textbf{Baselines and Ours}}
For each language group setting above, we conducted NMT experiments on both the multilingual training from scratch (\textbf{MLSC})~\cite{johnson-etal-2017-googles,aharoni-etal-2019-massively} and the mBART multilingual fine-tuning (\textbf{mBART FT})~\cite{DBLP:journals/corr/abs-2008-00401} as baselines. We applied our proposed word-level contrastive learning in both MLSC and mBART FT, and compared with another strong baseline, word alignment based joint NMT training (\textbf{+align})~\cite{garg-etal-2019-jointly}. 
For applying our method, we investigated the performance of joint training with word pairs extracted by both word2word (\textbf{+w2w}) and FastAlign (\textbf{+FA}). We omitted~\citet{lin-etal-2020-pre} as a baseline because their method can not be applied to mBART fine-tuning, and they used high-quality ground-truth dictionaries, which are unavailable for most languages pairs.

\noindent{\textbf{Implementation}}
We used \texttt{mBART-large} (mBART-25) for mBART FT and \texttt{transformer-base}~\cite{DBLP:conf/nips/VaswaniSPUJGKP17} for MLSC. See Appendix~\ref{app:hparams} for details.

\begin{figure}[t!]
\begin{minipage}[t]{\linewidth}
\begin{center}
\includegraphics[width=0.85\linewidth]{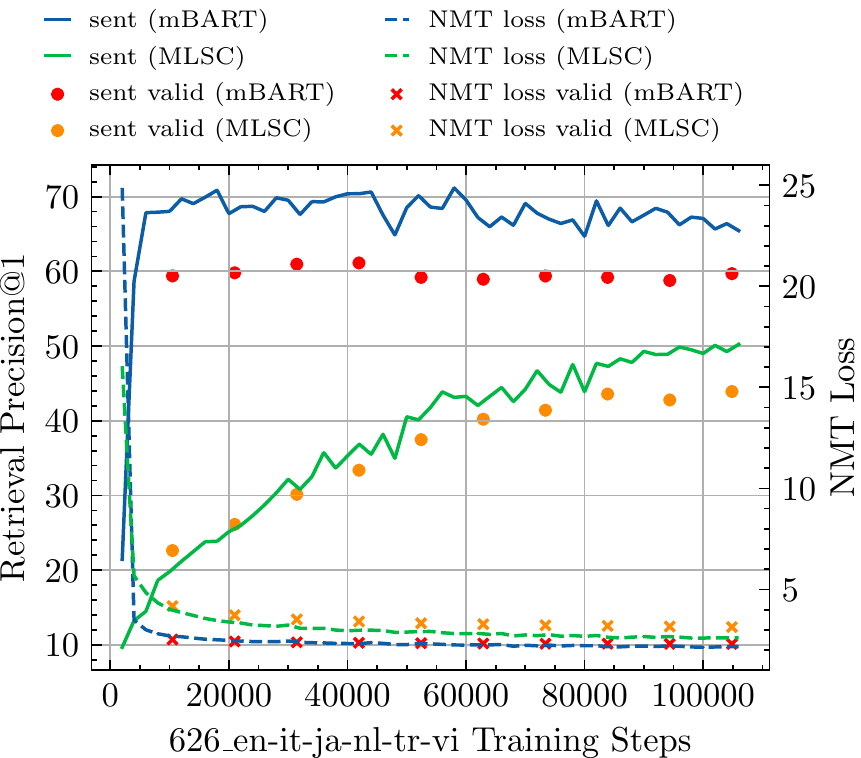}
\end{center}
\end{minipage}
\caption{\textbf{NMT loss, sentence retrieval P@1 of the encoder in MLSC and mBART FT.} The average of the contextual embeddings on top of the encoder is used as the sentence embedding. We report the average in-batch retrieval precision of both directions of each language pair.}
\label{encoder-alignment-p}
\end{figure}

\begin{figure*}[t]
\begin{minipage}[t]{0.33\textwidth}
\begin{center}
\includegraphics[width=0.95\linewidth]{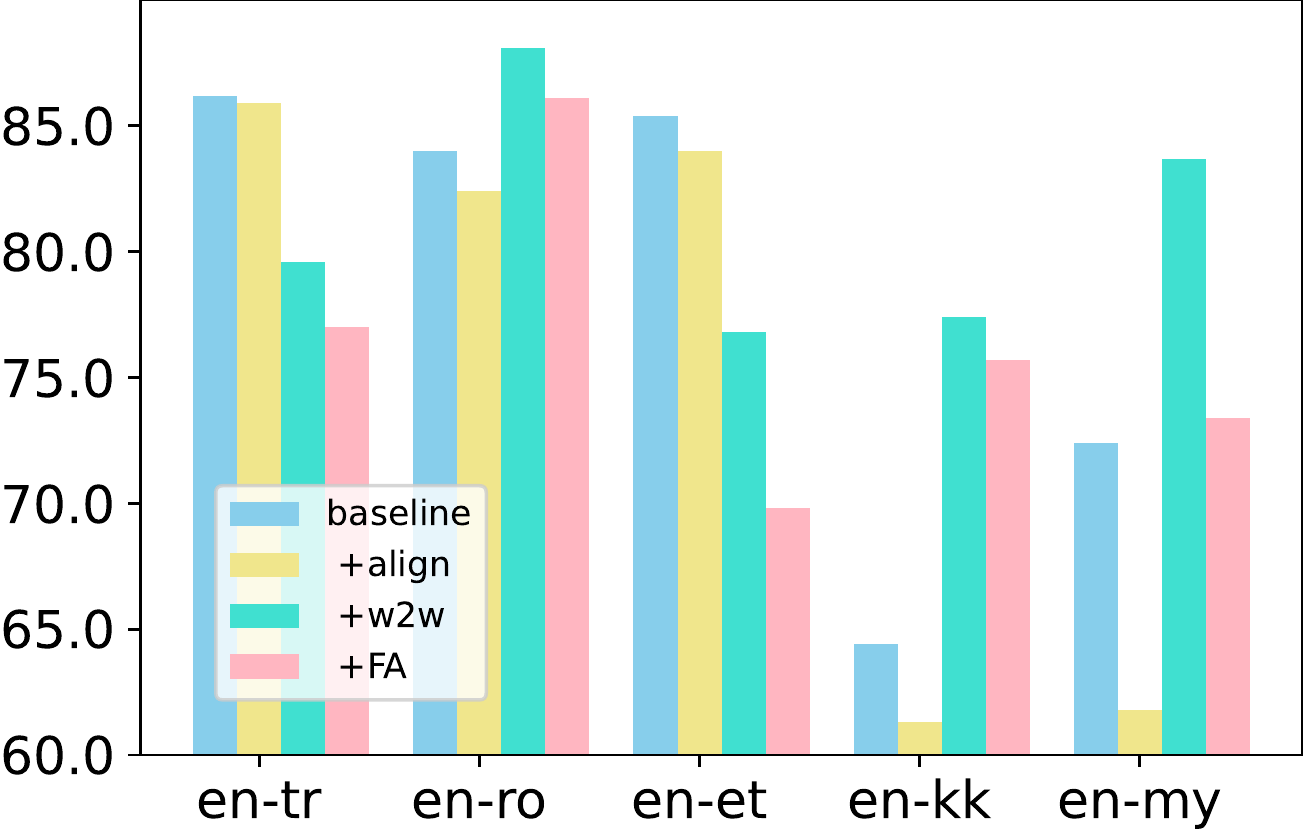}
\end{center}
\end{minipage}
\begin{minipage}[t]{0.33\textwidth}
\begin{center}
\includegraphics[width=0.95\linewidth]{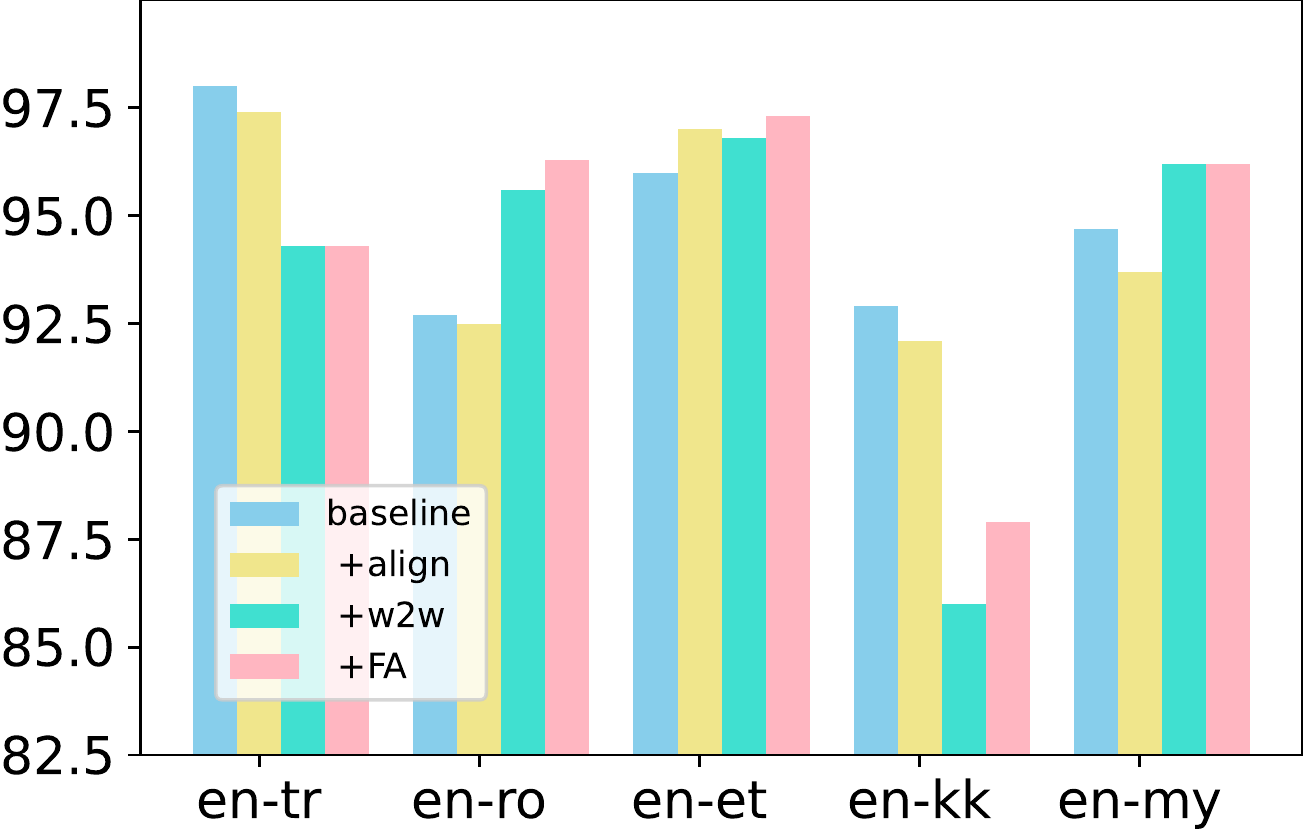}
\end{center}
\end{minipage}
\begin{minipage}[t]{0.33\textwidth}
\begin{center}
\includegraphics[width=0.95\linewidth]{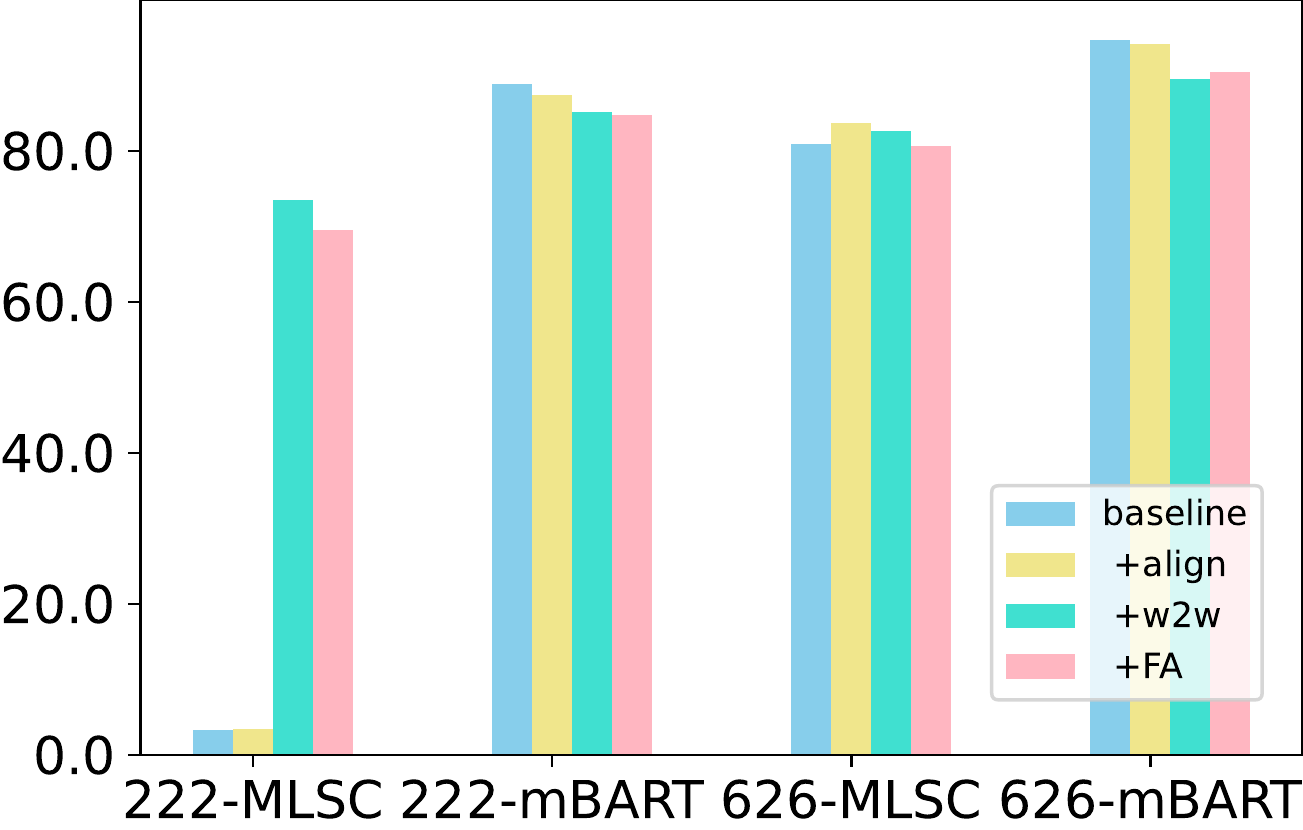}
\end{center}
\end{minipage}
\caption{\textbf{Sentence retrieval P@1 on the validation set for each language pair.} \textit{Left} and \textit{middle} are the results on 626\_en-tr-ro-et-my-kk MLSC and mBART FT, respectively. ``626'' in \textit{right} subfigure denote 626\_en-it-ja-nl-tr-vi. Refer to Appendix~\ref{sec:app4} for setup and results in details.}
\label{sr}
\end{figure*}

\section{Results and Analyses}
\label{sec:4}

\noindent{\textbf{BLEU Results}}
We report case-sensitive tokenized BLEU~\cite{papineni-etal-2002-bleu} results in Table~\ref{bleu1} and~\ref{bleu2}. In Table~\ref{bleu1}, we observe that with our proposed training objectives, BLEU scores are comparable in 222\_en-ja and 626\_en-it-ja-nl-tr-vi while they are slightly improved in 626\_en-tr-ro-et-my-kk. However, ``+align'' performs comparable or even worse compared with the baseline. Referring to Table~\ref{bleu2} for specific BLEUs on each language pair, we find that with our methods, translation performances are significantly improved for mBART FT while nontrivial improvements can merely be observed on en-ro and en-kk direction for MLSC. This indicates that NMT fine-tuning on monolingual pre-trained models (mBART) may benefit more from our proposed methods. 
Note that the BLEU improvements for MLSC are not significant, and we explain why this happens in the ``Word Retrieval P@1 is improved'' part.

\noindent{\textbf{Latent Encoder Alignment Property}}
We now inspect which aspect of alignment-based methods impacts the translation performance. Previous work~\cite{artetxe-schwenk-2019-massively} showed that the encoder of a strong multilingual NMT system is an ideal model for the bilingual sentence retrieval task. In addition,~\citet{DBLP:journals/corr/abs-1903-07091} introduced the correlation between the encoder-side sentence representation\footnote{Usually a pooled encoder output.} and the translation quality. Inspired by these, we speculate that alignment-based objectives affect sentence retrieval performance, which further impacts the translation quality. We train MLSC and mBART FT and report the sentence retrieval precision and NMT loss during the training. Results are reported in Figure~\ref{encoder-alignment-p}. We observe that the validation retrieval precision show similar trends as the NMT loss. This indicates that during many-to-many NMT training from scratch, encoder-side sentence-level retrieval precision is optimized along with the NMT loss. 




\noindent{\textbf{Sentence Retrieval P@1 Correlates with BLEU}}
According to the investigation of the encoder alignment property above, we verify the relationship between BLEU score and sentence retrieval precision on the validation set for each language pair. Results are shown in Figure~\ref{sr}. Cross-referencing the BLEU score in Table~\ref{bleu2}, we found that BLEU scores are improved when the encoder achieves gains on the sentence retrieval precision.\footnote{222\_en-ja MLSC setting can hardly learn a well-aligned encoder while our methods improve the encoder sentence-level alignment quality without sacrificing BLEU scores.} For example, we see increases of the retrieval P@1 on en-ro, en-et, and en-my on mBART FT (the middle of Figure~\ref{sr}) while BLEU scores are significantly improved on these three language pairs (Table~\ref{bleu2}). We further calculate the Pearson correlation coefficient between the BLEU changes and sentence retrieval P@1 changes for mBART+align, mBART+w2w, and mBART+FA in the 626\_en-tr-ro-et-my-kk setting. Results are 0.79, 0.93, 0.90, respectively, demonstrating a strong correlation between translation quality and sentence retrieval precision.

\begin{figure}[t!]
\begin{minipage}[t]{\linewidth}
\begin{center}
\includegraphics[width=0.9\linewidth]{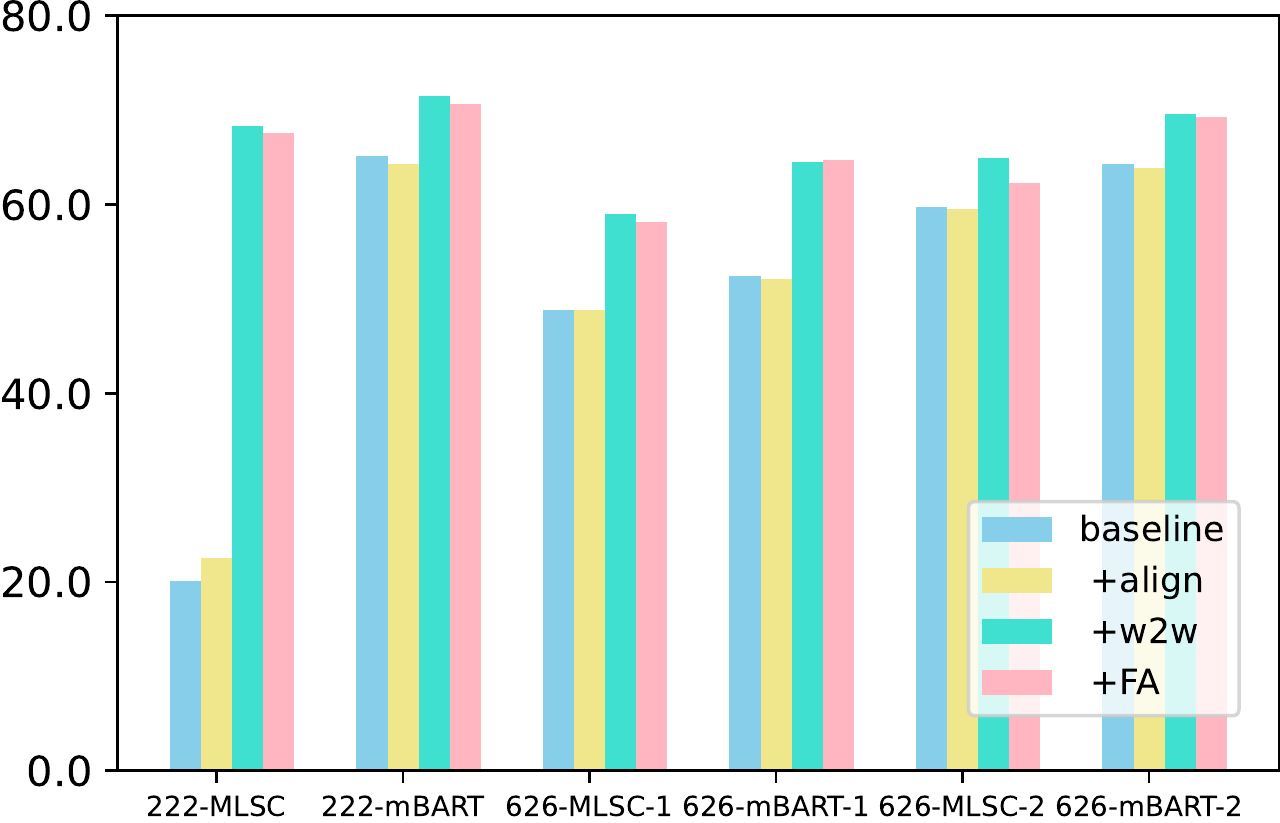}
\end{center}
\end{minipage}
\caption{\textbf{Average Word retrieval P@1 on the validation set for each language pair.} ``626-*-1'' and ``626-*-2'' indicate 626\_en-it-ja-nl-tr-vi and 626\_en-tr-ro-et-my-kk, respectively. Refer to Appendix~\ref{sec:app5} for setup and results in details.}
\label{word-retrieval}
\end{figure}

\noindent{\textbf{Word Retrieval P@1 is Improved}}
We probe the trained contextualized word representations on top of the encoder. As shown in Figure~\ref{word-retrieval}, we observe that the word retrieval precision is improved in all the settings. This demonstrates that the encoder parameters of the NMT system trained with our proposed objective are of a rather different distribution. By just changing the random seed, we can expect similar BLEU results, but we cannot obtain a better aligned encoder. However, the improvement of the word retrieval precision does not directly contribute to the translation quality, which we explain next.

\noindent{\textbf{Word-level Contrastive Objective and Sentence Retrieval P@1}} 
With the word-level contrastive objective, we observed significant BLEU score improvements on language pairs such as en-ro, en-et and en-my for mBART FT as presented in Table~\ref{bleu2}. However, noisy word pairs~\cite{pan-etal-2021-multilingual} extracted via word alignment toolkits leads to poor supervision signals for improving sentence retrieval P@1, which in turn prevents some language pairs such as en-kk from exhibiting BLEU improvements. We found that for en-kk, the numbers of extracted word pairs per sentence by word2word and FastAlign are 1.0 and 2.2, respectively. In contrast, these numbers are 4.2 and 20.7 for improved language pairs, calculated from Table~\ref{tab1}. Although better extracted word alignments for the word-level contrastive objective leads to BLEU improvements, its contribution towards improvements varies for MLSC and mBART FT, as shown in Table~\ref{bleu2}. We expect these findings to provide new perspectives for improving many-to-many NMT.

\noindent{\textbf{Sentence-level Contrastive Objective}}
We conducted the experiments for the sentence-level contrastive objective~\cite{pan-etal-2021-contrastive} on all two six-to-six settings and compared it against our proposed approach. The average BLEUs of our methods significantly outperform those of sentence-level contrastive objectives (see Table~\ref{bleu4} and~\ref{bleu5}), clearly showing the sentence-level objective's limitation. Moreover, we checked the sentence retrieval P@1 for~\citet{pan-etal-2021-contrastive} (Table~\ref{sr2} and~\ref{sr3}) and found that it correlates with BLEU changes, indicating that sentence-level contrastive objective is suboptimal for language pairs with decreased retrieval precision.\footnote{Note that the sentence-level contrastive objective incorporates sentences in multiple languages for contrastive loss. It does not necessarily improve the pair-wise retrieval precision.}


\section{Conclusion}
We proposed a word-level contrastive learning objective for many-to-many NMT. Experimental results showed that our proposed method leads to significantly better translation for several language pairs, which is then explained by analyses showing the relationship between BLEU scores and sentence retrieval performance of the NMT encoder. Future work can focus on: (1) further improving the encoder's retrieval performance in many-to-many NMT; (2) contrastive objective's feasibility in a massively multilingual scenario.

\section*{Ethical Considerations}
All the corpora we used in this paper are publicly available resources without the issue of the copyright. The technique this paper proposed is for NMT models, so it can not circumvent the issues that NMT models have. Since our automatically dictionaries are extracted from potentially biased data, the translations may also contain biases. However, we expect that these issues may be resolved by using unbiased data or the addition of debiasing objectives.


\section*{Acknowledgements}
This work was supported by Grant-in-Aid for Young Scientists \#19K20343, JSPS and Information/AI/Data Science Doctoral Fellowship of Kyoto University.

\bibliography{anthology,custom}
\bibliographystyle{acl_natbib}

\clearpage
\appendix

\begin{table}[t]
    \centering
    \resizebox{\linewidth}{!}{
    \begin{tabular}{llllr}
    \toprule
        La. pair & Train & Valid & Test & OD Size\\
         \hline
         en-et & WMT18 & WMT18 &	WMT18 & 10.7M \\
         en-it & IWSLT17 & IWSLT15 & IWSLT16 & 13.6M \\
         en-ja & IWSLT17 & IWSLT15 & IWSLT16 & 10.7M \\
         en-kk & WMT19 & WMT19 & WMT19 & 851k \\
         en-my & ALT & ALT & ALT & 446k \\
         en-nl & IWSLT17 & IWSLT15 & IWSLT16 & 12.7M \\
         en-ro & WMT16 & WWT16 & WMT16 & 11.0M \\
         en-tr & WMT17 & WWT16 & WMT16 & 11.1M \\
         en-vi & IWSLT15 & IWSLT13 & IWSLT14 & 11.9M \\
         \bottomrule
    \end{tabular}
    }
    \caption{\textbf{Dataset statistics for each language pair.} ``La. pair'' means language pair and ``OD Size'' denotes the number of the out-of-domain sentence pairs used for training FastAlign.}
    \label{dataset}
\end{table}

\begin{table}[t!]
    \centering
    \begin{tabular}{lrrrrrrrrrrr}
    \toprule
        Methods & en-ja & ja-en\\
        \toprule
        MLSC & 15.9 &	11.9\\
        \hline
        \ \ +align & \cellcolor{cyan!25}16.3 &	11.5\\
        \ \ +w2w (ours) & \cellcolor{cyan!25}16.0 &	11.7\\
        \ \ +FA (ours) & 15.6 &	11.0\\
        \toprule
        mBART FT & 19.8 &	18.0\\
        \hline
        \ \ +align & 19.6 &	17.5\\
        \ \ +w2w (ours) & 19.4 &	\cellcolor{cyan!25}18.2\\
        \ \ +FA (ours) & 19.5 &	17.8\\
        \bottomrule
    \end{tabular}
    \caption{\textbf{BLEU scores of 222\_en-ja system.} Significantly better scores are in cyan, and marginal improvements are in lightcyan. The significance test is done with~\citet{koehn-2004-statistical}.}
    \label{bleu3}
\end{table}

\begin{table}[t]
    \centering
    \begin{tabular}{lrrrrrrrrrrr}
    \toprule
        Methods & en-ja \\
        \toprule
        MLSC & 3.3 \\
        \hline
        \ \ +align & \textbf{3.5} \\
        \ \ +w2w (ours) & \textbf{73.5} \\
        \ \ +FA (ours) & \textbf{69.6} \\
        \toprule
        mBART FT & 88.9 \\
        \hline
        \ \ +align & 87.4 \\
        \ \ +w2w (ours) & 85.2 \\
        \ \ +FA (ours) & 84.8 \\
        \bottomrule
    \end{tabular}
    \caption{\textbf{Sentence retrieval P@1 on the validation set for 222\_en-ja.}}
    \label{sr1}
\end{table}

\begin{table}[t]
    \centering
    \begin{tabular}{lr}
    \toprule
        Methods & en-ja \\
        \toprule
        MLSC & 20.1 \\
        \hline
        \ \ +align & \textbf{22.5} \\
        \ \ +w2w (ours) & \textbf{68.3} \\
        \ \ +FA (ours) & \textbf{67.6} \\
        \toprule
        mBART FT & 65.2 \\
        \hline
        \ \ +align & 64.3 \\
        \ \ +w2w (ours) & \textbf{71.5} \\
        \ \ +FA (ours) & \textbf{70.7} \\
        \bottomrule
    \end{tabular}
    \caption{\textbf{Word retrieval P@1 on the validation set for 222\_en-ja.}}
    \label{wr1}
\end{table}

\begin{table*}[t!]
    \centering
    \begin{tabular}{lrrrrrrrrrrr}
    \toprule
        \multirow{2}{*}{Methods} & \multicolumn{2}{c}{en-ja} & \multicolumn{2}{c}{en-vi} & \multicolumn{2}{c}{en-it} & \multicolumn{2}{c}{en-nl} & \multicolumn{2}{c}{en-tr} & \multirow{2}{*}{Avg.}\\
        &$\rightarrow$&$\leftarrow$&$\rightarrow$&$\leftarrow$&$\rightarrow$&$\leftarrow$&$\rightarrow$&$\leftarrow$&$\rightarrow$&$\leftarrow$ & \\
        \toprule
        MLSC & 15.4 &	11.8 &	29.6 &	28.6 &	27.5 &	32.7 &	29.1 &	36.4 &	11.6 &	14.9 &	23.76\\
        \hline
        \ \ +align & 15.1 &	11.4 &	29.4 &	28.3 &	\cellcolor{cyan!25}27.7 &	\cellcolor{cyan!25}33.0 &	28.9 &	36.0 &	\cellcolor{cyan!25}11.8 &	\cellcolor{cyan!25}15.1 &	23.67\\
        \ \ +w2w (ours) & 15.3 &	11.6 &	\cellcolor{cyan!25}29.7 &	28.2 &	\cellcolor{cyan!25}27.6 &	32.4 &	28.6 &	35.8 &	10.8 &	14.4 &	23.44\\
        \ \ +FA (ours) & \cellcolor{cyan!25}15.5 &	11.6 &	29.6 &	28.0 &	\cellcolor{cyan!25}27.8 &	\cellcolor{cyan!80}33.2 &	29.1 &	35.9 &	11.2 &	14.9 &	23.68\\
        \hline
        \ \ +sent & 15.1	&11.6&	29.6&	28.3&	27.3&	32.7&	28.1	&\cellcolor{cyan!25}36.6&	11.3	&14.7	&23.53 \\
        \toprule
        mBART FT & 17.8 &	17.0 &	34.1 &	35.7 &	32.5 &	38.0 &	32.6 &	41.6 &	18.7 &	23.1 &	29.11\\
        \hline
        \ \ +align & 17.6 &	16.7 &	33.7 &	35.6 &	32.0 &	37.7 &	32.5 &	41.3 &	18.7 &	22.9 &	28.87\\
        \ \ +w2w (ours) & 17.6 &	\cellcolor{cyan!25}17.2 &	\cellcolor{cyan!25}34.2 &	35.7 &	32.5 &	\cellcolor{cyan!25}38.2 &	32.1 &	\cellcolor{cyan!25}41.7 &	18.7 &	22.9 &	29.08\\
        \ \ +FA (ours) & 17.5 &	\cellcolor{cyan!80}17.7 &	34.0 &	35.2 &	32.4 &	37.9 &	32.3 &	41.4 &	18.6 &	23.1 &	29.01\\
        \hline
        \ \ +sent & 17.8&	16.5	&33.7&	35.6	&32.2&	\cellcolor{cyan!25}38.1&	32.5&	41.2&	18.1&	22.9&	28.86 \\
        \bottomrule
    \end{tabular}
    \caption{\textbf{BLEU scores of 626\_en-it-ja-nl-tr-vi system.} Significantly better scores are in cyan, and marginal improvements are in lightcyan. The significance test is done with~\citet{koehn-2004-statistical}.}
    \label{bleu4}
\end{table*}

\begin{table*}[t]
    \centering
    \begin{tabular}{lrrrrrrrrrrr}
    \toprule
        \multirow{2}{*}{Methods} & \multicolumn{2}{c}{en-tr} & \multicolumn{2}{c}{en-ro} & \multicolumn{2}{c}{en-et} & \multicolumn{2}{c}{en-kk} & \multicolumn{2}{c}{en-my} & Avg. \\
        &$\rightarrow$&$\leftarrow$&$\rightarrow$&$\leftarrow$&$\rightarrow$&$\leftarrow$&$\rightarrow$&$\leftarrow$&$\rightarrow$&$\leftarrow$&\\
        \toprule
        MLSC & 9.3&	12.6&	25.0&	26.2&	10.8&	15.1& 0.5	&5.3	&15.1	&15.6&13.55\\
        \hline
        \ \ +align & 9.0&	12.4	&24.6&\cellcolor{cyan!25}	26.5&	10.7&	14.6&	0.4	&\cellcolor{cyan!25}5.4&	15.0&	15.3&	13.39\\
        \ \ +w2w (ours) &\cellcolor{cyan!25}9.4&	12.6&	24.8&	\cellcolor{cyan!80}26.8&	10.8&	15.1&0.5&	\cellcolor{cyan!80}5.8&	\cellcolor{cyan!25}15.2&	\cellcolor{cyan!25}15.9 	&13.69\\
        \ \ +FA (ours) & 9.1&	12.2&	24.8&	\cellcolor{cyan!80}26.7&	10.7&	14.8& 	0.3&	\cellcolor{cyan!25}5.6&	15.0&	15.6&13.48	\\
        \hline
        \ \ +sent & 8.7	&12.1&	24.5&	26.0&	10.4&	14.5	&0.4&	5.3	&13.8&	14.6	&13.03 \\
        \toprule
        mBART FT & 17.7&	22.2&	33.8&	37.1&	14.5&	24.3&	1.8&	14.1&	17.8&	23.1 & 20.64\\
        \hline
        \ \ +align & 17.5	&21.9&	33.8&	36.7&	\cellcolor{cyan!80}15.2	&24.3	&1.8&	14.0&	16.9&	22.1&	20.42\\
        \ \ +w2w (ours) &17.6&	22.2&	\cellcolor{cyan!25}34.2&	\cellcolor{cyan!25}37.5&	\cellcolor{cyan!80}15.0&	\cellcolor{cyan!80}25.0&1.2&	14.1&	\cellcolor{cyan!80}18.3&	\cellcolor{cyan!80}23.8 & 20.89\\
        \ \ +FA (ours) & 17.5&	22.2&	\cellcolor{cyan!80}34.3&	\cellcolor{cyan!25}37.5	&\cellcolor{cyan!25}14.9&	\cellcolor{cyan!80}25.1& 1.3&	\cellcolor{cyan!25}14.4&	\cellcolor{cyan!25}17.9&	\cellcolor{cyan!80}23.6 & 20.87\\
        \hline
        \ \ +sent &17.2&	22.1&	\cellcolor{cyan!25}34.2&	37.0&	14.2&	24.1&	1.6	&14.0&	17.7&	\cellcolor{cyan!25}23.4&	20.55 \\
        \bottomrule
    \end{tabular}
    \caption{\textbf{BLEU scores of 626\_en-tr-ro-et-my-kk system.} Significantly better scores are in cyan, and marginal improvements are in lightcyan. The significance test is done with~\citet{koehn-2004-statistical}.}
    \label{bleu5}
\end{table*}

\section{Tokenization Settings}
\label{sec:app1}

For Japanese, we use Jumanpp~\cite{morita-etal-2015-morphological,tolmachev-etal-2018-juman} for segmentation, and we follow the same settings as in mBART~\cite{liu-etal-2020-multilingual-denoising} for other languages: \texttt{myseg.py}~\cite{DBLP:journals/talip/DingAPNSUS20} is used for Burmese, Moses tokenization and special normalization is used for Romanian following~\cite{sennrich-etal-2016-edinburgh},\footnote{\url{https://github.com/rsennrich/wmt16-scripts}} and Moses tokenization for other languages.\footnote{\url{https://github.com/moses-smt/mosesdecoder/blob/master/scripts/tokenizer/tokenizer.perl}} Following mBART, we apply SentencePiece~\cite{kudo-richardson-2018-sentencepiece} to further segment sentences into subwords.\footnote{\url{https://github.com/google/sentencepiece}}

\section{Datasets and Alignment Extraction}
\label{sec:app2}
The datasets used for NMT training, validation and test are shown in Table~\ref{dataset}. For the word alignment extraction using FastAlign, we also use out-of-domain parallel corpora to train the FastAlign jointly, aiming to obtain word alignments with less noise. The out-of-domain corpora for all the language pairs contain Tatoeba, Europarl, GlobalVoices, NewsCommentary, OpenSubtitles, TED, WikiMatrix, QED, GNOME, bible-uedin, and ASPEC~\cite{nakazawa-etal-2016-aspec}. We collect them from the OPUS project~\cite{DBLP:journals/lre/Christodoulopoulos15} and WAT.\footnote{\url{https://lotus.kuee.kyoto-u.ac.jp/WAT/WAT2021/index.html}} The number of the out-of-domain parallel sentences for each language pair is shown in Table~\ref{dataset}.

\begin{table}[t]
    \centering
    \resizebox{\linewidth}{!}{
    \begin{tabular}{lrrrrrrrrrrr}
    \toprule
        Methods & en-ja & en-vi & en-it & en-nl & en-tr & Avg. \\
        \toprule
        MLSC & 52.7 &	84.6 &	91.0 &	85.7 &	89.7 &	80.9\\
        \hline
        \ \ +align & \textbf{53.5} &	82.8 &	\textbf{91.2} &	\textbf{86.4} &	88.9 &	80.6\\
        \ \ +w2w (ours) & \textbf{73.4} &	\textbf{85.7} &	\textbf{91.4} &	84.7 &	83.1 &	\textbf{83.7}\\
        \ \ +FA (ours) & \textbf{71.3} &	\textbf{84.9} &	\textbf{91.3} &	83.8 &	82.0 &	\textbf{82.7}\\
        \hline
        \ \ +sent & 87.2&	84.7&	91.1&	87.7&	86.6&	87.5\\
        \toprule
        mBART FT & 87.1 &	96.2 &	97.3 &	94.6 &	98.5 &	94.7\\
        \hline
        \ \ +align & 85.1 &	95.8 &	97.3 &	94.2 &	98.5 &	94.2\\
        \ \ +w2w (ours) & 81.6 &	91.4 &	94.7 &	90.8 &	89.6 &	89.6\\
        \ \ +FA (ours) & 82.6 &	92.3 &	95.0 &	91.7 &	90.4 &	90.4\\
        \hline
        \ \ +sent & 76.2&	88.3&	93.6&	88.7&	89.8&	87.3\\
        \bottomrule
    \end{tabular}
    }
    \caption{\textbf{Sentence retrieval P@1 on the validation set for 626\_en-it-ja-nl-tr-vi.}}
    \label{sr2}
\end{table}

\begin{table}[t]
    \centering
    \resizebox{\linewidth}{!}{
    \begin{tabular}{lrrrrrrrrrrr}
    \toprule
        Methods & en-tr & en-ro & en-et & en-kk & en-my & Avg. \\
        \toprule
        MLSC & 86.2 &	84.0 &	85.4 &	64.4 &	72.4 &	78.5\\
        \hline
        \ \ +align & 85.9 & 82.4 & 84.0 & 61.3 &	61.8 & 75.1\\
        \ \ +w2w (ours) & 79.6 &	\textbf{88.1} &	76.8 &	\textbf{77.4} &	\textbf{83.7} &	\textbf{81.1}\\
        \ \ +FA (ours) & 77.0 &	\textbf{86.1} &	69.8 &	\textbf{75.7} &	\textbf{73.4} &	76.4\\
        \hline
        \ \ +sent & 76.3	&77.6	&55.2&	63.8	&71.4&	68.9 \\
        \toprule
        mBART FT & 98.0 &	92.7 &	96.0 &	92.9 &	94.7 &	94.9\\
        \hline
        \ \ +align & 97.4&	92.5&	\textbf{97.0}	&92.1&	93.7&	94.5\\
        \ \ +w2w (ours) & 94.3 &	\textbf{95.6} &	\textbf{96.8} &	86.0 &	\textbf{96.2} &	93.8\\
        \ \ +FA (ours) & 94.3 &	\textbf{96.3} &	\textbf{97.3} &	87.9 &	\textbf{96.2} &	94.4\\
        \hline
        \ \ +sent & 94.6	&\textbf{97.3}&	95.4&	\textbf{93.1}	&\textbf{95.7}	&\textbf{95.2}\\
        \bottomrule
    \end{tabular}
    }
    \caption{\textbf{Sentence retrieval P@1 on the validation set for 626\_en-tr-ro-et-my-kk.}}
    \label{sr3}
\end{table}

\section{Implementation Details}
\label{app:hparams}

Following~\citet{DBLP:journals/corr/abs-2008-00401}, we set the oversampling temperature of 1.5 for all the settings. For MLSC, we set the dropout of 0.3 to avoid overfitting on small-scale training data. We used the batch size of 1,024 tokens for all the settings. For our word-level contrastive learning, we set the weight of 0.1, the temperature of 0.2, $d'$ of 128, and a smaller dropout of 0.2 because our proposed objective serves as a regularization part. We followed the hyperparameter setting of~\citet{garg-etal-2019-jointly} for word alignment-based joint NMT training. We used 8 NVIDIA A100 for mBART FT and 8 TITAN Xp for MLSC model training. The model is validated every 1000 steps for 222\_en-ja and 2000 steps for both two 626 settings. We do the early stopping if no improvement of the validation loss is observed for 8 checkpoints. The model with the best validation loss was used for evaluation.

\section{BLEU Scores}
\label{sec:app3}

We report all the BLEU results of 222\_en-ja, 626\_en-it-ja-nl-tr-vi, and 626\_en-tr-ro-et-my-kk in Table~\ref{bleu3},~\ref{bleu4} and~\ref{bleu5}, respectively.

\section{Sentence Retrieval Precision}
\label{sec:app4}

We report the sentence retrieval precision for all the systems in Tables~\ref{sr1},~\ref{sr2} and~\ref{sr3}. The sentence retrieval previsions are evaluated by using the validation dataset of each language pair. The mean pooled encoder output is used as the sentence embedding. We use cosine similarity to conduct the retrieval task, and report the average retrieval precision of both directions of each language pair.



\begin{table}[t]
    \centering
    \resizebox{\linewidth}{!}{
    \begin{tabular}{lrrrrrrrrrrr}
    \toprule
        Methods & en-ja & en-vi & en-it & en-nl & en-tr & Avg. \\
        \toprule
        MLSC & 61.8&	54.6&	42.8&	42.1&	42.7&	48.8\\
        \hline
        \ \ +align & \textbf{61.9}	&54.1&	\textbf{43.7}	&42.0&	42.3&	48.8\\
        \ \ +w2w (ours) & \textbf{64.0}	&\textbf{64.7}	&\textbf{55.8}&	\textbf{57.7}	&\textbf{52.8}&	\textbf{59.0}\\
        \ \ +FA (ours) & 58.2	&\textbf{65.2}&	\textbf{59.2}&	\textbf{60.1}&	\textbf{48.1}	&\textbf{58.2}\\
        \toprule
        mBART FT & 64.5	&57.2&	47.4&	45.9&	47.2&	52.4\\
        \hline
        \ \ +align & 64.0&	56.8&	47.3&	45.7&	46.8	&52.1\\
        \ \ +w2w (ours) & \textbf{71.3}&	\textbf{70.1}&	\textbf{60.6}&	\textbf{62.9}&	\textbf{57.8}&	\textbf{64.5}\\
        \ \ +FA (ours) & \textbf{68.6}&	\textbf{69.4}&	\textbf{63.2}	&\textbf{64.7}&	\textbf{57.4}&	\textbf{64.7}\\
        \bottomrule
    \end{tabular}
    }
    \caption{\textbf{Word retrieval P@1 on the validation set for 626\_en-it-ja-nl-tr-vi.}}
    \label{wr2}
\end{table}

\begin{table}[t]
    \centering
    \resizebox{\linewidth}{!}{
    \begin{tabular}{lrrrrrrrrrrr}
    \toprule
        Methods & en-tr & en-ro & en-et & en-kk & en-my & Avg. \\
        \toprule
        MLSC & 41.9&	63.2&	64.4&	63.4&	65.8&	59.7\\
        \hline
        \ \ +align & 40.9&	63.2&	63.9	&63.4&	\textbf{66.2}&	59.5\\
        \ \ +w2w (ours) & \textbf{50.1}&	\textbf{66.5}&	\textbf{67.6}	&\textbf{68.8}&	\textbf{71.3}&	\textbf{64.9}\\
        \ \ +FA (ours) & \textbf{47.2}&	\textbf{66.7}&	\textbf{65.7}&	\textbf{65.4}&	\textbf{66.3}&	\textbf{62.3}\\
        \toprule
        mBART FT & 46.8&	66.1&	68.0	&68.7&	71.7	&64.3\\
        \hline
        \ \ +align & 46.4&	65.9&	67.8&	68.5&	71.1	&63.9\\
        \ \ +w2w (ours) & \textbf{55.6}	&\textbf{70.3}&	\textbf{72.8}&	\textbf{74.7}&	\textbf{74.4}&	\textbf{69.6}\\
        \ \ +FA (ours) & \textbf{55.3}	&\textbf{70.1}&	\textbf{73.0}&	\textbf{74.0}&	\textbf{74.0}&	\textbf{69.3}\\
        \bottomrule
    \end{tabular}
    }
    \caption{\textbf{Word retrieval P@1 on the validation set for 626\_en-tr-ro-et-my-kk.}}
    \label{wr3}
\end{table}

\section{Word Retrieval Precision}
\label{sec:app5}
We report the word retrieval precision for all the systems in Tables~\ref{wr1},~\ref{wr2}, and~\ref{wr3}. The word retrieval precision are computed by using the validation dataset and the word2word alignments on it. The mean pooled encoder output on corresponding positions is used as the contextualized word embedding. We use cosine similarity to implement the retrieval for word pairs in a batch, and present the average in-batch retrieval precision of both directions of each language pair. Batch size is set as 512 tokens.

\end{document}